# Deep Gaussian Covariance Network

**Kevin Cremanns & Dirk Roos**
Institute of Modelling and High-Performance Computing
University of Applied Sciences Niederrhein, Krefeld, Germany
Kevin.cremanns@hs-niederrhein.de
Dirk.roos@hs-niederrhein.de

## Abstract

The correlation length-scale $\theta$ next to the noise variance $\sigma^2$ are the most used hyperparameters for the Gaussian processes $\mathcal{GP}$. Typically stationary covariance functions $k(\boldsymbol{x}_i, \boldsymbol{x}_j)$ are used, which are only dependent on the distances between input points $\tau = ||\boldsymbol{x}_i - \boldsymbol{x}_j||$ and thus invariant to the translations in the input space $X$. The optimization of the hyperparameters is commonly done by maximizing the log marginal likelihood $\log p(\boldsymbol{y}|X, \theta, \sigma^2)$. This works quite well, if $\boldsymbol{\tau} \sim \mathcal{U}(0, max(\tau))$, since $\theta$ fits for every subspace in $X$. In the case of $\boldsymbol{\tau} \not\sim \mathcal{U}(0, max(\tau))$ like in a locally adapted or even sparse $X$, the prediction $\hat{\boldsymbol{y}}_{\mathcal{GP}}(\boldsymbol{x}_*)$ of a test point $\boldsymbol{x}_*$ can be worse dependent of its position. A possible solution to this, is the usage of a non-stationary covariance function $k(\tau_*, \theta_*)$, where $\theta_*$ is calculated by a deep neural network $dnn(\boldsymbol{x}_*)$. So that the correlation length scales $\theta_*$ and also possibly the noise variance $\sigma_*^2$ are dependent on the test point $\boldsymbol{x}_*$. Furthermore, different types of covariance functions are trained simultaneously, so that the $\mathcal{GP}$ prediction is an additive overlay of different covariance matrices $\hat{\boldsymbol{y}}_{\mathcal{GP}}(X_*) = K_1(X, X_*, \boldsymbol{\theta_1}(X_*), \boldsymbol{\sigma}_1^2(X_*)) + ... + K_n(X, X_*, \boldsymbol{\theta_n}(X_*), \boldsymbol{\sigma}_n^2(X_*))$. The right covariance functions combination and its hyperparameters $\boldsymbol{\theta}(X_*), \boldsymbol{\sigma}^2(X_*)$ are learned by the deep neural network. Additional, the $\mathcal{GP}$ will be able to be trained by batches or online and so it can handle arbitrarily large data sets. We call this framework Deep Gaussian Covariance Network (DGCN). There are also further extensions to this framework possible, for example sequentially dependent problems like time series or the local mixture of experts. The basic framework and some extension possibilities will be presented in this work. Moreover, a comparison to some recent state of the art surrogate model methods will be performed, also for a time dependent problem.

## 1 Introduction

The main purpose of the supervised learning is to build a model $\hat{f} : \mathbb{R}^{d_{in}} \to \mathbb{R}^{d_{out}}$ for an unknown relationship of the data $Y = f(X)$. This data can come for example from measurements, experiments or computer simulations, which may be very time or cost expensive. Therefore unnecessarily big training data are tried to be avoided. A smart method of adaption is used in order to create as less training samples as possible like presented in [1], [2] or [3]. For example in optimization task or reliability analysis these methods are used to increase the model prediction in the area of interest. Since these adaptation methods always follow some objective, it can happen, that the density of the training samples $X$ are locally much higher then in the overall design space. If the data comes from measurements it is also often the case, that the training data is not uniformly distributed $X \not\sim \mathcal{U}$ and sparse. A further problematic scenario is the data sets with discontinuities, like in the field of robotic movements.

The Gaussian process model ($\mathcal{GP}$) [4] is a commonly used supervised learning algorithm. Like a lot of other surrogate model methods, there exists hyperparameters, which need to be adjusted to fit the model to the training data $X$. In the case of the Gaussian process the correlation length-scale $\theta$ next to the noise variance $\sigma^2$ are the most used hyperparameters. The correlation length $\theta$ describes basically how much the known observations $y$ influence the prediction $\hat{y}_{\mathcal{GP}}$, based on the distance $\tau = ||\boldsymbol{x}_i - \boldsymbol{x}_j||$. If $\boldsymbol{\theta}$ is a vector with the same dimension as the input parameters $n_v$, then each element of $\boldsymbol{\theta}$ describe the influence along each dimension axis and can also be used for automatic relevance determination (ARD) [5]. The inverse of the length-scale determines, how relevant an input dimension is and if it gets very large it can even diminish the influence on the covariance of a specific input parameter. After the explanation of the meaning of $\theta$, it should be clear, why stationary correlation functions might not be as good as a non-stationary in the case of $\boldsymbol{\tau} \not\sim \mathcal{U}(0, max(\boldsymbol{\tau}))$.

There exist a variety of related works on the topic of non-stationary $\mathcal{GP}$ correlation functions. A lot of the publications use an input transformation like via a neural network, which results in a mapping of $X$ in a higher or lower dimensional space $\mathbb{R}^{d_{in}} \to \mathbb{R}^{d_{in}\pm}$ ([6], [7], [8]). Other authors proposed extending the $\mathcal{GP}$ directly with input-dependent parameters. These parameters are treated as separate Gaussian processes and inferred jointly with the unknown function ([9], [10], [11]). Non-stationarities can also be included by the use of non-stationary variants of kernel functions, which was first introduced by [12] and extended by [13] with Marcov chain Monte Carlo (MCMC). One of the most recent works is [14], where for the first time a non-stationary and heteroscedastic $\mathcal{GP}$ regression framework was presented, in which the three main components noise variance, signal variance and length-scale can be simultaneously input-dependent, with direct $\mathcal{GP}$ priors. They used Hamiltonian Monte Carlo instead of introducing variational or expectation propagation approximations, and place $\mathcal{GP}$ priors for each of the parameters. This resulted in 3 $\mathcal{GP}$ priors, where they used the same correlation function. A similar work was presented in [15], where also a Gaussian prior was placed to estimate the local smoothness and so the local length-scale. The usage of one or multiple $\mathcal{GP}$ to estimate the input dependent hyperparameter of the main $\mathcal{GP}$, makes the model training expensive if the data set are too large.

In this paper a novel approach is presented, which uses a deep neural network ($dnn$) to learn the input dependent non-stationary hyperparameters $\theta, \sigma^2$ of the $\mathcal{GP}$ together with the combination of various different covariance functions to increase the prediction quality. It would also be possible to extend this approach to the signal variance. However like [14] already mentioned, these parameter have in most cases a small influence on the prediction quality of the model and will therefore be neglected in the further investigation. The presented framework is highly flexible for further extension like e.g. local mixture of experts or manifold mapping of the design space like proposed in [8]. Additionally, we will show its capabilities for sequential dependent problems (e.g. time series). Further more it brings the possibilities to perform batch or online learning to Gaussian processes, which makes it possible to use arbitrarily large data sets $X$ for training. Big data sets are a common problem for $\mathcal{GP}$ because of the distance calculation and the inversion of the covariance matrix during the training and the prediction. To optimize our model, we use neural network training algorithms like ADAM [16]. The non-stationary length-scales $\boldsymbol{\theta}$ (with one $\theta$ per input variable) can also be used for local sensitivity estimates. So they can indicate the importance of variables dependent of space or even time. The framework is implemented in Keras [17] with the Theano [18] backend, which makes it quite fast as we will demonstrate. Moreover, since $\sigma^2$ will also be estimated dependent on the input point, it will be very useful in the case of mixed low and high fidelity data or if not all observations in the training data $X$ have the same noise level.

In the following section 2 the novel approach, called Deep Gaussian Covariance Network ($\mathcal{DGCN}$), will be explained by firstly giving a short revision about the Gaussian process method and neural networks in 2.1 and 2.2. In 2.3 there will be the description of the deep learning approach for non-stationary $\mathcal{GP}$ hyperparameter estimation. Afterward, there will also be shown some extension possibilities of the presented framework in 2.4. In section 3 the approach will be tested against recent state of the art methods and a conclusion will be made in section 4.

## 2 Deep Gaussian Covariance Network

### 2.1 Gaussian Process

The Gaussian process [4] is a nonparametric Bayesian regression method, which is very powerful for a lot of supervised problems [19], [20]. As already mentioned the choice of the covariance function and its hyperparameters are the main drivers of the prediction quality of the model. One common choice for the correlation function is the squared exponential function:

$$k(\boldsymbol{\tau}, \boldsymbol{\theta}) = \exp(-0.5\boldsymbol{\theta}\boldsymbol{\tau}^2) \qquad (1)$$

Each correlation function depends on the correlation lengths $\boldsymbol{\theta}$ (with one correlation length per input parameter in our case) and the distances $\boldsymbol{\tau}$ between the training and test points $X, X_*$. Although the squared exponential can be applied to a great range of problems, it might be the wrong choices for all problems. There exists further covariance functions like the Matérn functions [21] or rational quadratic function. There exists also the possibility to build new covariance function out of existing ones by sum or multiply different covariance matrices ([22], [23]). This allows to build complex covariance matrices without the decision for only one maybe suitable correlation function. Further it gives more hyperparameter to estimate dependent on the number of used covariance matrices there exists for each of them a set of $\boldsymbol{\theta}, \sigma^2$.

The $\mathcal{GP}$ equation to estimate the prediction mean $\hat{\boldsymbol{y}}_{\mathcal{GP}}$ and the variance $\mathbb{V}[\hat{\boldsymbol{y}}_{\mathcal{GP}}]$ is given by:

$$\hat{\boldsymbol{y}}_{\mathcal{GP}} = K_*(X, X_*, \boldsymbol{\theta})^T (K(X, X, \boldsymbol{\theta}) + \sigma^2 I)^{-1} \boldsymbol{y} \qquad (2)$$

$$\mathbb{V}[\hat{\boldsymbol{y}}_{\mathcal{GP}}] = K_*(X_*, X_*, \boldsymbol{\theta}) - K_*^T(X, X_*, \boldsymbol{\theta})(K(X, X, \boldsymbol{\theta}) + \sigma^2 I)^{-1} K_*^T(X, X_*, \boldsymbol{\theta}) \qquad (3)$$

where $K_*$ denotes the covariance matrix dependent on the points to predict $X_*$ and the training points $X$ with size $N \times N_*$ and $K$ denotes the covariance matrix dependent on $X, X$ with size $N \times N$. $N, N_*$ denotes the number of training and test points. $\boldsymbol{y}$ describe the response values.

The covariance matrix values are further dependent on the chosen covariance function $k(\boldsymbol{\tau}, \boldsymbol{\theta})$, which are dependent on the euclidean distances $\boldsymbol{\tau} = \|X_* - X\|$ and the correlation length-scales $\boldsymbol{\theta}$. The noise variance $\sigma^2$ is added to the diagonal of $K(X, X, \boldsymbol{\theta})$ and also a free hyperparameter.

One of the very useful properties of $\mathcal{GP}$ is the estimation of the prediction variance $\mathbb{V}[\hat{\boldsymbol{y}}_{\mathcal{GP}}]$. Since it can be used to estimate the prediction confidence intervals via e.g. the t-student distribution (it yields the assumption for the Gaussian process that the prediction error is $\sim \mathcal{N}(0, \sigma^2)$):

$$CI_{\hat{\boldsymbol{y}}_{\mathcal{GP}}} = \hat{\boldsymbol{y}}_{\mathcal{GP}} \pm \mathcal{T}\left(\frac{\sqrt{\mathbb{V}[\hat{\boldsymbol{y}}_{\mathcal{GP}}]}}{\sqrt{N}}, 1 - \alpha\right) \qquad (4)$$

The optimization of the hyperparameters will usually be performed by maximizing the marginal log likelihood under the assumption $\boldsymbol{y} \sim \mathcal{N}(0, K + \boldsymbol{\sigma}^2 I)$:

$$\log p\left(\boldsymbol{y} | X, \boldsymbol{\theta}, \sigma^2\right) = -0.5 \boldsymbol{y}^T (K(X, X, \boldsymbol{\theta}) + \sigma^2 I)^{-1} \boldsymbol{y} \\ - 0.5 \log(\det(K(X, X, \boldsymbol{\theta}) + \sigma^2 I)) - 0.5 N \log(2\pi) \qquad (5)$$

The gradient w.r.t. its hyperparameters $\boldsymbol{\theta}$ and $\sigma^2$, where we treat $\sigma^2$ as a part of $\boldsymbol{\theta}$ (under the assumption of a Gaussian likelihood [24]), can be used for optimization:

$$\begin{aligned} \frac{\partial}{\partial \theta_j} \log p(\boldsymbol{y}|X, \boldsymbol{\theta}) &= \frac{\partial \log p\left(\boldsymbol{y}|X, \boldsymbol{\theta}, \sigma^2\right)}{\partial K(X, X, \boldsymbol{\theta})} \frac{\partial K(X, X, \boldsymbol{\theta})}{\partial \boldsymbol{\theta}} \\ &= 0.5 \boldsymbol{y}^T K(X, X, \boldsymbol{\theta})^{-1} \frac{\partial}{\partial \theta_j} K(X, X, \boldsymbol{\theta})^{-1} \boldsymbol{y} \\ &- 0.5 tr\left(K(X, X, \boldsymbol{\theta})^{-1} \frac{\partial K(X, X, \boldsymbol{\theta})}{\theta_j}\right) \end{aligned} \qquad (6)$$

$tr$ means the trace of the resulting matrix.

## 2.2 Neural networks

A deep learning neural network $dnn$ can learn any function with a large enough number of hidden layers and neurons [25]. This makes it a powerful tool for any kind of supervised regression or classification problem. The simplest way to explain how they work is to begin with a single neuron network, as shown in Fig. 1. The prediction equation for one point $x$ is:

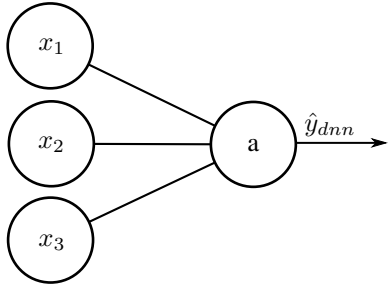

Figure 1: Single neuron neural network.

$$\hat{y}_{dnn}(\boldsymbol{x}) = a\left(\sum_{i=1}^{n_v} w_i x_i + b\right) \tag{7}$$

where $a()$ describes the activation function, e.g. the sigmoid function:

$$f(x) = \frac{1}{1 + exp(-x)} \tag{8}$$

$w_i$ is the corresponding weight and $b$ is the bias term, which represent the two typical hyperparameters of a neural network. If we extend our model to $n_n^l$ neurons in the hidden layer $l$ and a further output layer with 1 neuron as the output, so in total the number of layers is $n_l = 3$, the prediction yields:

$$\hat{y}_{dnn}(\boldsymbol{x}) = \sum_{j=1}^{n_n^{(3)}} a_j^{(3)} \left( \sum_{i=1}^{n_n^{(2)}} W_{ij} \sum_{j=1}^{n_n^{(2)}} a_j^{(2)} \left( \sum_{i=1}^{n_v^{(1)}} W_{ij} x_i + b_j^{(1)} \right) + b_j^{(2)} \right) \tag{9}$$

$i$ denotes the i-th unit of layer $l$ and $j$ denotes the unit of layer $l+1$. The upper script denotes layer $l$. So the hyperparameters are $W, \boldsymbol{b} = \left(W^{(1)}, b^{(1)}, ..., W^{(n_l)}, b^{(n_l)}\right)$ containing the weights and biases of all $n_l$ layers. This can be extended to arbitrary number of hidden layers.

The cost function for regression is typically the one-half squared error:

$$J(W, \boldsymbol{b}, \boldsymbol{X}, \boldsymbol{y}) = 0.5 ||\hat{\boldsymbol{y}}_{dnn}(W, \boldsymbol{b}, X) - \boldsymbol{y}||^2 \tag{10}$$

The partial derivatives of $J(W, \boldsymbol{b}, \boldsymbol{X}, \boldsymbol{y})$ w.r.t. its hyperparameters $W, \boldsymbol{b}$ can be used can be used for optimization via backpropagation and gradient descent [26], which is a very basic approach. One iteration of gradient descent updates the parameters $W, \boldsymbol{b}$ as follows:

$$W_{i,j}^{(l)} = W_{i,j}^{(l)} - \alpha \frac{\partial}{\partial W_{ij}^{(l)}} J(W, \boldsymbol{b}) \tag{11}$$

$$b_i^{(l)} = b_i^{(l)} - \alpha \frac{\partial}{\partial b_i^{(l)}} J(W, \boldsymbol{b}) \tag{12}$$

with the learning rate $\alpha$ to control the gradients. The partial derivatives can be obtained via:

$$\frac{\partial}{\partial W_{ij}^{(l)}} J(W, \boldsymbol{b}) = \frac{1}{N_b} \sum_{i=1}^{N_b} \frac{\partial}{\partial W_{i,j}^{(l)}} J(W, \boldsymbol{b}, \boldsymbol{x}_i, y_i) \quad (13)$$

$$\frac{\partial}{\partial b_i^{(l)}} J(W, \boldsymbol{b}) = \frac{1}{N_b} \sum_{i=1}^{N_b} \frac{\partial}{\partial b_i^{(l)}} J(W, \boldsymbol{b}, \boldsymbol{x}_i, y_i) \quad (14)$$

$N_b$ indicates the batch size with $1 \leq N_b \leq N$. $dnn$ are usually trained with stochastic algorithms like ADAM [16], which are also based on backpropagation. They perform better especially for large network topologies, since there exist a lot of local minimas.

Further extension to $dnn$ for sequential dependent problems like time series, are recurrent neural networks [27] and LSTM networks [28], which use previous steps to predict future steps.

Some advantages of $dnn$ are the easy computation, which makes them useful for big Data applications. In addition to the use of flexible training batch sizes, it can be easily used for distributed learning. Drawbacks are the manual choice of the network topology and the type of activation functions. Further, $dnn$ tend to over-fit the data, if no regularization methods are used, like dropout layers [29] or add additional noise to the data during training [30]. Compared to $\mathcal{GP}$ they tend to have a lower prognosis quality especially for small number of trainings samples $N$ ($N < 150$) ([31], [32]).

### 2.3 Deep learning for non-stationary GP hyperparameters

After giving a short explanation of the used methods, we will now explain how to combine a deep neural network to use it for non-stationary $\mathcal{GP}$ hyperparameter learning. Fig. 2 gives an overview how the framework works. The first part represents an arbitrary deep neural network $dnn$, which has got the training points $X$ as its input and the $\mathcal{GP}$ hyperparameters $\Theta, \sigma^2$ as its output. We use the notation $\Theta$ at this point because it is a matrix of size $N_b \times n_v$. So a $\boldsymbol{\theta}$ for each input point $\boldsymbol{x}$ with $n_v$ elements to reduce the influence of unimportant parameters. $\boldsymbol{\sigma}^2$ is now a vector because it is also dependent on the input point $\boldsymbol{x}$ with $N_b$ elements.

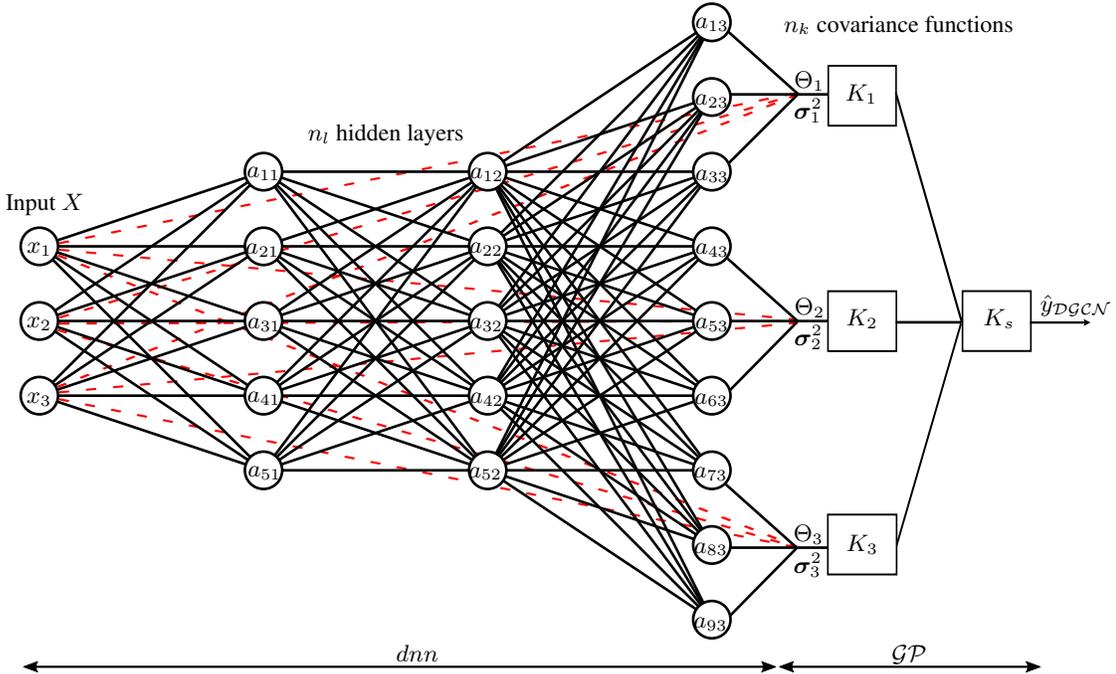

Figure 2: Schematic overview of the deep Gaussian covariance network.

The $\mathcal{GP}$ uses as its input also the training points $X$ and additional the output of the $dnn$; $\Theta, \boldsymbol{\sigma}^2$ and the observations $\boldsymbol{y}$. Its output will be the final prediction $\hat{\boldsymbol{y}}_{\mathcal{DGCN}}$ of the test points $X_*$. As already mentioned we want to use $n_k$ different correlation functions $k_i()$ and use the sum of the resulting covariance matrices as a new covariance matrix $K_s = K_1 + ... + K_{n_k}$. This extends the size of the $dnn$ output $\Theta$ to $N_b \times (n_v \times n_k)$.

Furthermore, $\Theta$ needs to be incorporated into the covariance functions $k_i$. Since the number of rows in $\Theta$ are equal to $N_b$ it is not possible to use the correlation lengths anymore like shown exemplary in Eq. 1, since $N_b$ is not equal to the dimension of $\boldsymbol{\tau}$. Therefore we need to reformulate the covariance functions, for example in the case for the squared exponential to:

$$k_i(X, X_*, \boldsymbol{\Theta}_i, \boldsymbol{\Theta}_{i*}) = \exp(-0.5||\boldsymbol{\Theta}_i X - \boldsymbol{\Theta}_{i*} X_*||^2) \tag{15}$$

$\boldsymbol{\Theta}_{i*}$ indicates the estimated correlation lengths from the $dnn$ for the test points $X_*$ with size $N_* \times n_v$. This can be done for all kind of covariance functions, which increases the calculation of $\boldsymbol{\tau}$ from 1 to $n_{ep} \times n_k$ ($n_{ep}$ number of epochs) during the training. One epoch is completed if all data points are used for the training once in the case of batch training.

Also Eq. 2 and 3 can be rewritten to:

$$\hat{\boldsymbol{y}}_{\mathcal{DGCN}} = K_{s*}(X, X_*, \Theta, \Theta_*)^T (K_s(X, X, \Theta, \Theta) + \boldsymbol{\sigma}^2 I)^{-1} \boldsymbol{y} \tag{16}$$

$$\mathbb{V}[\hat{\boldsymbol{y}}_{\mathcal{DGCN}}] = K_{s*}(X_*, X_*, \Theta_*, \Theta_*) - K_{s*}^T(X, X_*, \Theta, \Theta_*) \\ (K_s(X, X, \Theta, \Theta) + \boldsymbol{\sigma}^2 I)^{-1} K_{s*}^T(X, X_*, \Theta, \Theta_*) \tag{17}$$

For the training the marginal log likelihood can still be used from the Eq. 5 depending on the $dnn$ hyperparameters:

$$\log p(\boldsymbol{y}|X, W, \boldsymbol{b}) = -0.5 \boldsymbol{y}^T (K_s(X, X, \Theta(W, \boldsymbol{b}), \Theta(W, \boldsymbol{b})) + \boldsymbol{\sigma}^2 I)^{-1} \boldsymbol{y} \\ - 0.5 \log(\det(K_s(X, X, \Theta(W, \boldsymbol{b}), \Theta(W, \boldsymbol{b})) + \boldsymbol{\sigma}^2 I)) - 0.5 N \log(2\pi) \tag{18}$$

For the gradients w.r.t. the hyperparameters of the deep learning network, the chain rule can be used together with the Eq. 6, 13, 14 and 18:

$$\frac{\partial \log p(\boldsymbol{y}|X, W, \boldsymbol{b})}{\partial W} = \frac{\partial \log p(\boldsymbol{y}|X, \Theta)}{\partial K_s(X, X, \Theta(W, \boldsymbol{b}), \Theta(W, \boldsymbol{b}))} \\ \frac{\partial K_s(X, X, \Theta(W, \boldsymbol{b}), \Theta(W, \boldsymbol{b}))}{\partial \Theta(W, \boldsymbol{b})} \frac{\partial \Theta(W, \boldsymbol{b})}{\partial W} \tag{19}$$

$$\frac{\partial \log p(\boldsymbol{y}|X, W, \boldsymbol{b})}{\partial \boldsymbol{b}} = \frac{\partial \log p(\boldsymbol{y}|X, \Theta)}{\partial K_s(X, X, \Theta(W, \boldsymbol{b}), \Theta(W, \boldsymbol{b}))} \\ \frac{\partial K_s(X, X, \Theta(W, \boldsymbol{b}), \Theta(W, \boldsymbol{b}))}{\partial \Theta(W, \boldsymbol{b})} \frac{\partial \Theta(W, \boldsymbol{b})}{\partial \boldsymbol{b}} \tag{20}$$

Eq. 4 to calculate the confidence intervals of the prediction of $\hat{\boldsymbol{y}}_{\mathcal{DGCN}}$ is also still valid. As for the optimization algorithm, we suggest stochastic algorithms like ADAM [16] or even the Nesterov momentum version Nadam [33]. It is also recommended to use regularization methods like drop layers or Gaussian noise for the $dnn$ part, since it tends to overfit very fast. Further, separate networks are used for the $\Theta$ and $\boldsymbol{\sigma}^2$ hyperparameters with different activation functions in order to decouple the $dnn$ hyperparameters training of length-scales and noise variance.

As mentioned before, batch training should be used, in order to train arbitrarily large data sets $X, Y$. Although it is a common approach for the $dnn$ part to train with batches, it is not possible without the modification to perform batch training with the $\mathcal{GP}$ part. The $\mathcal{GP}$ needs for the prediction (Eq. 2) typically the whole data set $X$ and also the training part is performed with the whole data set $X, Y$. The training process might be possible without modifications but without a non-stationary input dependent training of the hyperparameters $\Theta, \boldsymbol{\sigma}^2$, like proposed in this work, the hyperparameters are always fitted to the last batch of the optimization. The next problem will be the prediction, since the distances between the test points $X_*$ and $X$ are needed. If the whole data set is not used, there needs to be a selection of the appropriate points near $X_*$.

To solve this problem, a nearest neighbor algorithm ([34]) is used, where the $N_b$ nearest points $X_{np}$ are used to predict a point $\boldsymbol{x}_*$. Therefore Eq. 16 and Eq. 17 need to be reformulated to:

$$\hat{\boldsymbol{y}}_{\mathcal{DGCN}} = K_{s*}(X_{np}, X_*, \Theta, \Theta_*)^T (K_s(X_{np}, X_{np}, \Theta, \Theta) + \boldsymbol{\sigma}^2 I)^{-1} \boldsymbol{y} \tag{21}$$

$$\begin{aligned}\mathbb{V}[\hat{\boldsymbol{y}}_{\mathcal{DGCN}}] =& K_{s*}(X_*, X_*, \Theta_*, \Theta_*) - K_{s*}^T(X_{np}, X_*, \Theta, \Theta_*) \\ & (K_s(X_{np}, X_{np}, \Theta, \Theta) + \boldsymbol{\sigma}^2 I)^{-1} K_{s*}^T(X_{np}, X_*, \Theta, \Theta_*)\end{aligned} \tag{22}$$

The idea behind this is, that typically for important parameters, the correlation length $\theta$ is only influenced by a small number of observed points in the region near the point to predict. If the parameter is unimportant, then it does not matter how many points lie around the point to predict regarding this specific parameter. This is only possible because of the non-stationary input dependent estimation of $\Theta$ and $\boldsymbol{\sigma}^2$. We will show an example of batch training in comparison to non-batch training in section 3. With the opportunity to use batch training for the proposed method, a later update for new training points of an existing model, e.g. online learning, is easily possible. Only the $W, \boldsymbol{b}$ of the $dnn$ part need to be updated, which is a common for neural networks.

### 2.4 Extensions

Before we show experiments with the new approach, this section will show two extension possibilities to this framework, which might be useful in some applications. In addition to the proposed method, feature mapping shown in [8] can also be connected with the presented framework. The Input can be mapped into a higher or lower dimensionality depending on the nodes in the output layer. This output will be the input for the presented framework. The feature mapping will be learned along with the hyperparameters of the $\mathcal{DGCN}$.

#### 2.4.1 Sequential dependent output

Recurrent neural networks or long short-term memory (LSTM) networks are common approaches to build models for sequential dependent outputs as mentioned in section 2.2. One advantage of such networks is the internal use of additional hyperparameters to train the sequential dependencies. An easy approach to emulate this behavior is to create artificial input parameters, which represents the input or output of previous sequential steps. For example for a time series the last two output results $y_{t-1}, y_{t-2}$ can be used as additional input parameters, with their own $\theta$ and $\sigma^2$. Therefore an additional pre-processing step of the training data is necessary, which shifts the data appropriately, as exemplary shown in Tab. 1. Section 3 will show also some comparison of the recurrent networks to the presented framework. By the definition of additional input parameters, the $dnn$ part will learn the sequential behavior over time similar to a recurrent network but the approximation is still based on the $\mathcal{GP}$ part. Also this time dependency is non-stationary dependent on $X$, which might result in a big improvement compared to stationary time dependency.

#### 2.4.2 Neural network based mixture of local experts

A further extension would be the inclusion of further models, e.g. a polynomial or even a physical based model, ($\hat{y}_{PM}$) in the manner like mixture of local experts work ([35], [36]). An additional network would be conducted to the presented approach, which controls the local mixture depending on the input points $X$ to archive this. Fig. 3 shows a exemplary framework for this. Then the $\mathcal{DGCN}$ learns its hyperparameters along with the output of existing models, in order to minimize the overall uncertainty of the connected models.

| t | $y_t$ |     | t | $x_{t-2}$ | $x_{t-1}$ | $y_t$ | $y_{t+1}$ | $y_{t+2}$ |
|---|---|---|---|---|---|---|---|---|
| 1 | 2 |   | 1 | - | - | - | - | - |
| 2 | 3 |   | 2 | - | - | - | - | - |
| 3 | 1 |   | 3 | 2 | 3 | 1 | 6 | 7 |
| 4 | 6 | → | 4 | 3 | 1 | 6 | 7 | 3 |
| 5 | 7 |   | 5 | 1 | 6 | 7 | 3 | 9 |
| 6 | 3 |   | 6 | 5 | 7 | 3 | 9 | 1 |
| 7 | 9 |   | 7 | - | - | - | - | - |
| 8 | 1 |   | 8 | - | - | - | - | - |

Table 1: Shifted time series for supervised learning

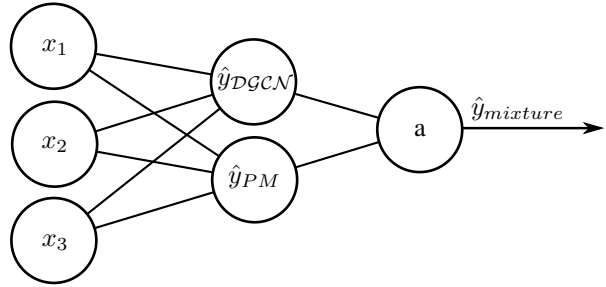

Figure 3: Neural network based local mixture of experts.

## 3 Experiments

For the following tests, the $dnn$ part of the $\mathcal{DGCN}$ is a deep learning network with 5 layers including input and output layer. The number of neurons per layer are $n_v \times 20 \times 20 \times 20 \times (n_v \times n_k)$, with $n_k = 5$. The activation functions are from layer 2 to 4: sigmoid, sigmoid, rectified linear unit and a linear output. The used modified correlation functions are:

- absolute and squared exponential:

$$k_{squared}(X, X_*, \mathbf{\Theta}_i, \mathbf{\Theta}_{i*}) = \exp\left(-0.5||\mathbf{\Theta}_i X - \mathbf{\Theta}_{i*} X_*||^2\right) \quad (23)$$

$$k_{abs}(X, X_*, \mathbf{\Theta}_i, \mathbf{\Theta}_{i*}) = \exp\left(-||\mathbf{\Theta}_i X - \mathbf{\Theta}_{i*} X_*||\right) \quad (24)$$

- the one and two times differentiable Matérn function:

$$k_{mat1}(X, X_*, \mathbf{\Theta}_i, \mathbf{\Theta}_{i*}) = \left(1 + \sqrt{3}||\mathbf{\Theta}_i X - \mathbf{\Theta}_{i*} X_*||\right) \exp\left(-\sqrt{3}||\mathbf{\Theta}_i X - \mathbf{\Theta}_{i*} X_*||\right) \quad (25)$$

$$k_{mat2}(X, X_*, \mathbf{\Theta}_i, \mathbf{\Theta}_{i*}) = \left(1 + \sqrt{5}||\mathbf{\Theta}_i X - \mathbf{\Theta}_{i*} X_*|| + \frac{5}{3}||\mathbf{\Theta}_i X - \mathbf{\Theta}_{i*} X_*||^2\right) \\ \exp\left(-\sqrt{5}||\mathbf{\Theta}_i X - \mathbf{\Theta}_{i*} X_*||\right) \quad (26)$$

- rational quadratic function:

$$k_{qr}(X, X_*, \mathbf{\Theta}_i, \mathbf{\Theta}_{i*}) = (1 + 0.25||\mathbf{\Theta}_i X - \mathbf{\Theta}_{i*} X_*||)^{-2} \quad (27)$$

Additionally, drop and noise layers are used in the training process. Further more a convergence criteria is used for early stopping.

### 3.1 Training time comparison

As a first experiment, the training speed should be compared for the batch training with $N_b = N$ and $N_b = 200$. Therefore different training sample sizes are used from 50 up to 1.638.400 points. The batch training with $N_b = N$ will only range up from 50 to 12.800 samples, since the memory demand exceed 16 GB (64-bit). The batch training with $N_b = 200$ will start from 400 training samples. Furthermore, the number of training epochs is fixed to 100, which is a sufficient number of training iterations to get a reasonable good model in most cases. Too much iterations would also lead to a over fitted model. Generally, a convergence criteria is used, which might also converge before 100 epochs. The test is performed on an Intel Core i7 3770 with 2 cores at 3.50 GHz and also on a Nvidia Quadro 4000 graphic card with 256 CUDA cores (7 years old, modern GPU have 3840 CUDA cores). As already mentioned in section 1, the implementation is based on Keras with Theano as its backend. The result are shown in Fig. 4. It is shown, that the usage of lower batch sizes $N_b$ are faster to calculate at a specific point over setting $N_b = N$. Even more important is that the lower batch size enables the $\mathcal{GP}$ part of the $\mathcal{DGCN}$ to handle any size of training samples $N$. The graphic card provide a speed up of factor 10. Modern graphic card would provide a even higher factor.

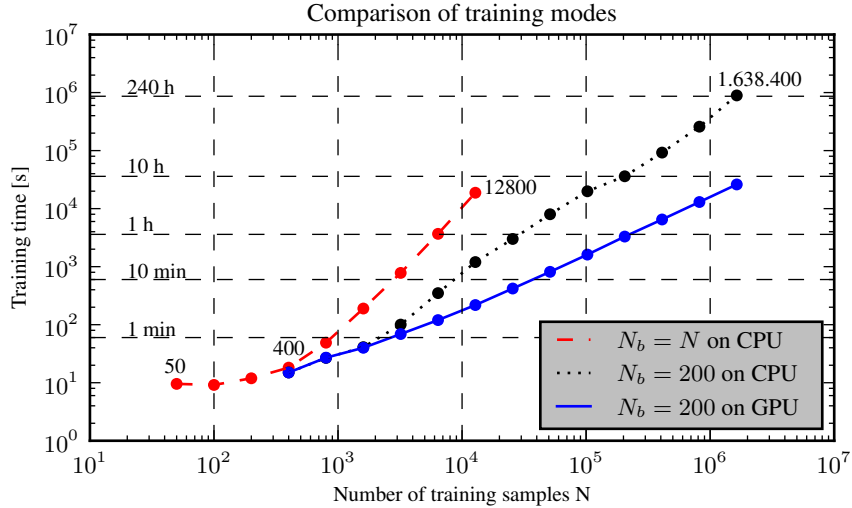

Figure 4: Training time comparison for different batch sizes $N_b$ on CPU and GPU.

### 3.2 Time series forecasting

The second experiment will be using the presented framework for a time dependent problem. Therefore we use the CATS benchmark [37], which was a competition for time series prediction. In the CATS benchmark, the goal was the prediction of 100 missing values of a 5000 point data set, see Fig. 5. These missing values were divided in 5 blocks, where the missing elements are:

- elements 981 to 1.000
- elements 1.981 to 2.000
- elements 2.981 to 3.000
- elements 3.981 to 4.000
- elements 4.981 to 5.000

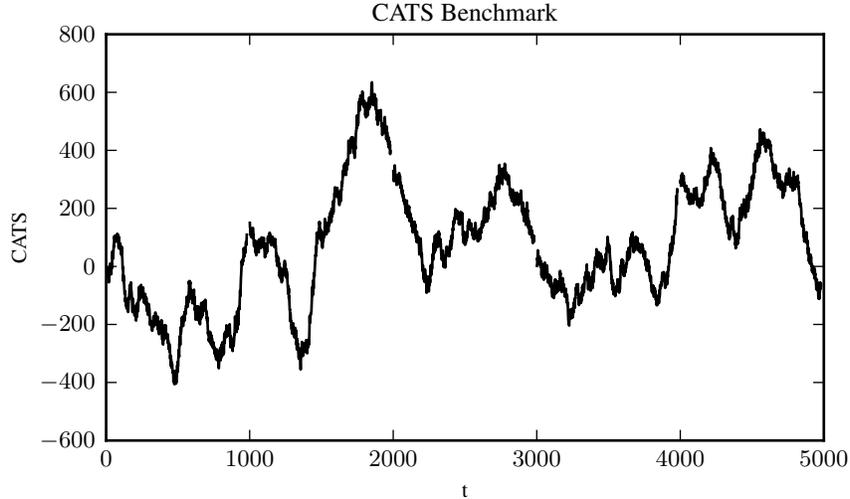

Figure 5: CATS benchmark 4900 data points and 100 missing points.

The competition objective was to minimize the mean square error $E_1$, which is calculated by:

$$E_1 = \frac{\sum_{t=981}^{1000}(y_t - \hat{y}_t)^2}{100} + \frac{\sum_{t=1981}^{2000}(y_t - \hat{y}_t)^2}{100} + \frac{\sum_{t=2981}^{3000}(y_t - \hat{y}_t)^2}{100} \\ + \frac{\sum_{t=3981}^{4000}(y_t - \hat{y}_t)^2}{100} + \frac{\sum_{t=4981}^{5000}(y_t - \hat{y}_t)^2}{100} \tag{28}$$

There was also a second criteria $E_2$, where only the first 80 of the 100 missing values are considered, for those who worked with the data before and after the missing values. This we do not consider here, since it was not the official main objective.

We build for this task 5 models, one for each block, where each of the models use a different number of previous time steps, in order to increase the maximum likelihood. The models are trained in such a way, that the $t - N_t$ previous time steps ($N_t$ denotes the number of previous used time steps) are used to predict the actual time step $t$. So only the data before the missing values are used for training to predict the last 20 elements of each block.

Overall 24 methods were submitted for this competition, with $E_1$ ranging from 408 to 1714. In Tab. 2, our result of $E_1 = 368$ is shown compared to only the first 3 of the competition. As shown our approach gets a much better result compared for example to deep recurrent neural networks [38].

| Author | $E_1$ | Model |
|--------|-------|-------|
| -      | 368   | Deep Gaussian covariance network |
| [39]   | 408   | Kalman Smoother |
| [38]   | 441   | Recurrent Neural Networks |
| [40]   | 502   | Competitive Associative Net |

Table 2: Results for the CATS benchmark

### 3.3 Regression example

The third experiment will show the capabilities for regression task of the presented framework. Therefore we use the often used Boston housing example [41], which summarize the median house prices in Boston metropolitan area. The data set consists of 506 data points with 13 input variables and 1 output. We compare our results to 5 different publications ([42], [43], [44], [45], [46]), that also used this example as a benchmark for their methods. Further results can also be found in [47].

For the first comparison to [42], where also non-stationary Gaussian processes were used., we follow the procedure to estimate the root mean squared error (RMSE) via 10-fold cross-validation. This is repeated 20 times, where the data is shuffled randomly for the cross-validation estimation. The data in this publication was transformed by taking the logarithm of the output to estimate the error.

The second comparison is made to [43]. This work uses also Gaussian process with a Student-t likelihood and a mixture of local experts to provide a more robust approach. In this case the first 455 samples were used for training and the last 51 samples for testing. The process was repeated 25 times. Also the data was transformed through normalization. Since the results are only visually shown and not the exact values, there is no exact comparison possible. We compare once again the RMSE. The visual results show a value of 0.44 as the lowest number on the y-axis. Where the exact value is $> 0.44$ and $< 0.46$.

The third comparison is made to [44]. Similar to the first comparison a Gaussian process is used with a Student-t likelihood to get a more robust result. Once again the data is normalized and the RMSE is estimated via 10-fold cross-validation. This time no repetition was made in this work. Anyway we show the results of 20 repetitions similar to the first comparison. Since we do not know, if this value was picked as the best result of multiple tries, it is difficult to compare the results but in our results the highest RMSE out of 20 repetitions is still a bit lower then the compared method value.

The fourth work we compare our results to is [45]. They use also Gaussian processes, with a special approach to choose the covariance function independently of the basis. They use also 10-fold cross-validation to estimate the mean squared error (MSE). We take the square root of their results to fit it to the other results. Also no repetitions were made. The output data was not transformed this time.

The last and fifth comparison is made to [46]. They use a very new and interesting approach called Deep Gaussian Processes [48] using stochastic expectation propagation and probabilistic Backpropagation. In this approach Gaussian processes are used as activation functions of a deep neural network. Similar to the first comparison 20 repetitions of a 10-fold cross-validation is used to estimate the RMSE. Instead of the min max value the standard deviation is given.

The results are summarized in Tab. 3. Overall to this 5 compared works, the $\mathcal{DGCN}$ was able to get better results in all cases. One thing to point out here is the fact that all comparison were done under the same settings given in section 3. Most of the compared methods show different setups, where we use the best results of. This shows that the presented approach is an easy to use method without the need of finding the right covariance functions of the $\mathcal{GP}$ or the right net topology of the $dnn$. Even if the kind of problem switches from time series forecasting to a regression task.

| Method | RMSE method | RMSE $\mathcal{DGCN}$ |
| --- | --- | --- |
| [42] | 0.1321 / 0.1346 / 0.1389 | 0.1205 / 0.1228 / 0.1244 |
| [43] | >0.44 / >0.44 / <0.46 | 0.401 / 0.408 / 0.412 |
| [44] | - / 0.287 / - | 0.259 / 0.2707 / 0.285 |
| [45] | - / 2.89 / - | 2.38 / 2.40 / 2.43 |
| [46] | - / 2.47±0.49 / - | - / 2.40±0.061 / - |

Table 3: Comparison of the Boston housing example

Since the Deep Gaussian Processes $\mathcal{DGP}$ [48] is a recent promising topic in the machine learning community, we want to use it for further comparison of other regression tasks. The work that is used for the comparison [46] includes further regression examples taken from the UCI database [49], for which we also test the presented framework. For all examples shown in Tab. 4 20 repetitions of 10-fold cross-validation were used to estimate the mean and standard deviation of the RMSE. The results were rounded to two decimals in their work. The $\mathcal{DGCN}$ was better in 4 of 5 examples compared to the $\mathcal{DGP}$. We excluded the classification examples or the examples, where no comparison is possible due their rounding to two decimals. For the examples kin8nm and power, batch training with batch size $N_b = 200$ were used.

| Example | $N$ | $n_v$ | $\mathcal{DGCN}$ | $\mathcal{DGP}$ |
|---------|-----|-------|------------------|-----------------|
| concrete | 1030 | 8 | **3.67 ± 0.06** | 5.21 ± 0.9 |
| energy 1 | 768 | 8 | **0.37 ± 0.01** | 0.48 ± 0.05 |
| energy 2 | 768 | 8 | **0.54 ± 0.01** | 1.37 ± 0.23 |
| kin8nm | 8192 | 8 | 0.06 ± 0.00 | **0.02 ± 0.00** |
| power | 9568 | 4 | **2.91 ± 0.00** | 2.95 ± 0.3 |

Table 4: Comparison of deep Gaussian process vs. deep Gaussian covariance network mean and standard deviation RMSE values for UCI database examples.

## 4 Conclusion

In this work a new framework, called deep Gaussian Covariance Network $\mathcal{DGCN}$, was presented. This approach uses a deep learning neural network to learn the non-stationary hyperparameters $\Theta, \boldsymbol{\sigma}^2$ of the Gaussian process dependent on the input points $X$. Furthermore, it learns the hyperparameters for multiple covariance functions simultaneously to further improve the prognosis quality and to avoid the manual choice of the appropriate covariance functions. It could be shown that this framework is easily extendable also for sequential dependent problems like time series, by using previous input and output steps as additional input, since then the model has hyperparameters describing the sequential dependencies. The predictive power of this approach could be shown for a time series benchmark and multiple regression examples, where this framework was in all cases except one better then the compared methods. It could also be shown that this approach is capable to work with arbitrary large data sets in a reasonable time. What was not mentioned in this work is the possibility of the grid search for the $dnn$ topology and the activation functions to further improve the predictive accuracy. We wanted to show, that with the same net topology several examples work very well, even if the problem changes from a time series problem to an ordinary regression task. So that one of the most difficult tasks by using deep learning neural networks can be neglected. The complexity of finding a network, which is capable to predict the right hyperparameters for the $\mathcal{GP}$ part is much easier then to find one for regression problem itself. Additionally, this framework can also be used for classification tasks, by using the appropriate likelihood function [4]. Further interesting experiments would be, to build models with low and high fidelity data, since the noise variance is also learnt depending on $X$ or to use this method for sampling adaption strategies.

## Acknowledgment

The authors would like to express their thanks to the German Federal Ministry for Economic Affairs and Energy, which supported this project, support identification no.: 03ET7071D.